# Autonomous Navigation of Unmanned Vehicle Through Deep Reinforcement Learning

Letian Xu[1],[*]
Independent Researcher
lottehsu@gmail.com

Jiabei Liu[2]
Northeastern University
USA
liu.jiabe@northeastern.edu

Haopeng Zhao[3]
New York University
USA
hz2151@nyu.edu

Tianyao Zheng[4]
Independent Researcher
tian971227@gmail.com

Tongzhou Jiang[5]
Independent Researcher
tojiang0111@gmail.com

Lipeng Liu[6]
Peking University
Beijing, China
lipeng.liu@pku.edu.cn

*Abstract*—This paper explores the method of achieving autonomous navigation of unmanned vehicles through Deep Reinforcement Learning (DRL). The focus is on using the Deep Deterministic Policy Gradient (DDPG) algorithm to address issues in high-dimensional continuous action spaces. The paper details the model of a Ackermann robot and the structure and application of the DDPG algorithm. Experiments were conducted in a simulation environment to verify the feasibility of the improved algorithm. The results demonstrate that the DDPG algorithm outperforms traditional Deep Q-Network (DQN) and Double Deep Q-Network (DDQN) algorithms in path planning tasks.

## I. INTRODUCTION

The field of autonomous navigation for unmanned vehicles has seen significant advancements in recent years, driven by the rapid development of deep reinforcement learning (DRL) techniques. DRL, with its ability to learn optimal policies through interaction with the environment, has become a powerful tool for enabling unmanned vehicles to navigate complex and dynamic environments autonomously.

Reference [1] introduces the concept of Learning from Teaching Regularization (LTR), which enhances the generalization capabilities of DRL models by ensuring that the learned correlations are easy to imitate. This method improves the robustness and adaptability of UVs in varied and unpredictable environments [2].

Reference [3] presents a deep learning-powered method for estimating the extrinsic parameters of unmanned surface vehicles. By leveraging convolutional neural networks (CNNs), this approach accurately determines the parameters necessary for precise navigation, thereby enhancing the overall operational efficiency of UVs in aquatic environments.

Reference [4] discusses the transition from auxiliary equipment to collaborative robot swarms in manufacturing, highlighting the importance of collaborative robots in modern industrial environments. Using DRL in this context enables robots to autonomously navigate and operate in dynamic environments, thereby improving efficiency and reducing human intervention [5].

Reference [6] introduces an edge-assisted epipolar transformer for industrial scene reconstruction. This innovative approach leverages edge computing to improve the accuracy and efficiency of scene reconstruction tasks [7].

Reference [8] developed a TD3-based collision-free motion planning method for robot navigation. The method improves the safety and reliability of autonomous navigation by enabling the robot to learn from past experience and predict future states to avoid obstacles. The TD3 algorithm is able to handle continuous action spaces and improve sampling efficiency, making it particularly suitable for real-time navigation in unpredictable environments.

Reference [9] introduced a resilient parachute positioning mechanism designed for detecting and tracking during the autonomous aerial refueling process of UAVs. The system embodies the practical application of DRL in improving the accuracy and reliability of complex tasks. By focusing on real-time detection and tracking, this research improves the capabilities of UAVs in critical operations and ensures a more efficient and accurate autonomous refueling process [10].

Reference [11] introduced MaxK-GNN, a GPU kernel design aimed at accelerating the training of Graph Neural Networks (GNNs). While this work primarily focuses on enhancing computational efficiency, the implications for autonomous navigation are significant.

Reference [12] enhanced visual-inertial odometry performance by integrating a GRU with a Kalman Filter[13][14]. This combination significantly improves the accuracy and robustness of odometry in complex environments, which is crucial for the reliable navigation of unmanned

vehicles[15][16]. The integration of these advanced techniques ensures that vehicles can maintain precise localization[17] and navigation even in challenging and dynamic settings [18][19].

This paper proposes the DDPG algorithm, which addresses the problem of optimizing path planning for autonomous navigation. This algorithm directly generates usable local trajectories by leveraging the continuous action space of DDPG. It effectively learns to navigate complex and dynamic environments, enabling the vehicle to avoid obstacles and achieve efficient path planning through end-to-end reinforcement learning.-

## II. METHODOLOGY

Reinforcement learning is a powerful machine learning paradigm for training agents to make decisions by interacting with an environment. Traditional RL algorithms like Q-learning and SARSA are effective in discrete action spaces, but their applicability to continuous action spaces is limited. To address this challenge, the Deep Deterministic Policy Gradient (DDPG) algorithm was proposed, combining the strengths of Deep Q-Network (DQN) and Policy Gradient methods. DDPG is an off-policy algorithm designed for environments with high-dimensional continuous action spaces.

### A. Mobile Robot Model

The robot autonomous navigation studied in this paper is based on a four-wheeled vehicle employing an Ackermann steering mechanism. This steering model is widely used in car-like robots and vehicles to achieve efficient turning and maneuverability by controlling the angles of the front wheels.

An Ackermann steering mechanism is characterized by its front steering wheels and rear driving wheels. The front wheels are steered at angles $\delta_f$ and $\delta_r$ relative to the vehicle's heading, while the rear wheels provide the driving force. This configuration allows the vehicle to turn smoothly and efficiently, maintaining the wheels' perpendicular orientation to the path's tangent.

The position and orientation of the vehicle can be described by its coordinates $(x, y)$ in the plane and the orientation angle $\theta$. Given the vehicle's velocity $v$ and the steering angle $\delta$, its position and orientation over time can be described. The position and orientation at time $t$ are denoted as $(x(t), y(t), \theta(t))$. The kinematic equations of the vehicle are:

$$\dot{x} = v\cos(\theta)$$
$$\dot{y} = v\sin(\theta) \quad (1)$$
$$\dot{\theta} = \frac{v}{L}\tan(\delta)$$

$L$ is the distance between the front and rear axles, and $\delta$ is the steering angle of the front wheels.

### B. DDPG algorithm

The deep deterministic policy gradient algorithm (DDPG) used in this paper is a policy learning method that outputs continuous actions. It is derived from the deterministic policy gradient (DPG) algorithm. It draws on the advantages of the single-step update of the Actor-Critic policy gradient and combines the experience replay and target network technology of the deep Q network (DQN) to improve the convergence of the Actor-Critic method. The DDPG algorithm consists of a policy network and a target network. DDPG uses a deterministic strategy to select actions, so the output is not the probability of the behavior, but the specific behavior. $\theta^\mu$ is the parameter of the policy network, $a_t$ is the action, and $s_t$ is the state. The target network will fix the parameters in the network within a certain period of time, thereby eliminating the model oscillation caused by the same parameters between the current network and the target network. The DDPG algorithm has a strong deep neural network fitting and generalization ability, as well as the advantage of processing continuous action space, and continuously trains and adjusts the neural network parameters by learning the optimal action strategy in the current state.

Applying this method to the path planning process of the mobile robot can enable the mobile robot to output more continuous actions and reduce decision errors during movement. In the process of path planning, the mobile robot obtains the state $S$ according to the surrounding environment information and its own state data, and the actor current network outputs the action a of the mobile robot according to $S$. After the mobile robot performs an action, it will get a reward from the environment. The current critic network outputs the Q value as the evaluation of the action according to $S$ and $a$, and continuously adjusts its value function. The actor current network continuously improves the action strategy according to the Q value. The target network of the actor and critic is mainly used in the subsequent update process. The structure of the DDPG algorithm is shown in Figure 1.

Actor network consists of a policy network that selects actions based on the current state and a target policy network that is a slowly updated copy of the policy network. The policy network outputs a specific action $\mu(s_i | \theta^\mu)$ given the state $s_t$.

Critic network evaluates the action taken by the actor by estimating the value of the state-action pair $Q(s_t, a_t | \theta^Q)$. It includes a target evaluation network, which is also a slowly updated copy, ensuring stability in training.

The target networks are updated using soft updates to ensure stability:

$$\begin{cases} \theta^{\mu'} \leftarrow \tau\theta^\mu + (1-\tau)\theta^{\mu'} \\ \theta^{Q'} \leftarrow \tau\theta^Q + (1-\tau)\theta^{Q'} \end{cases} \quad (2)$$

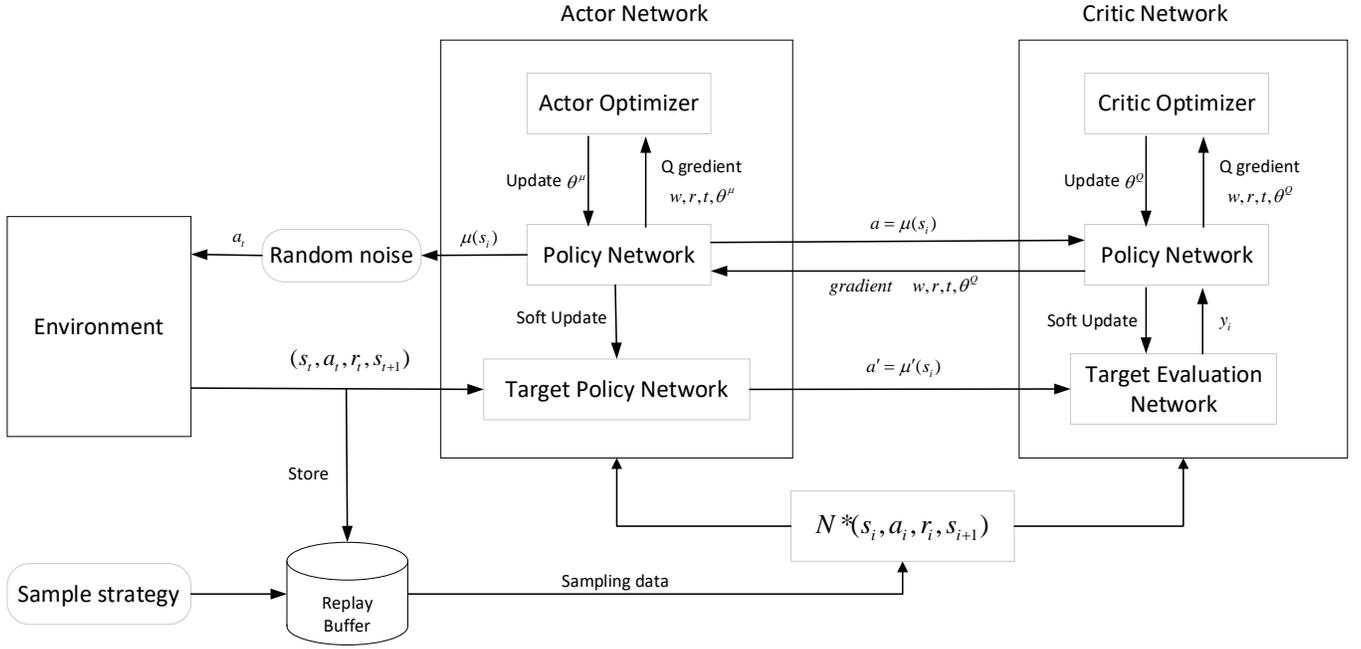

Fig 1. Structure of DDPG algorithm

## III. EXPERIMENTS

This experiment is conducted in a simulation environment to explore and learn, aiming to verify the feasibility of the improved algorithm. The main purpose of the training is to enable the robot to avoid collisions in an environment with obstacles and reach the target point through a smooth and shortest path. The experimental training rules are as follows: the initial point of each round is fixed, the target point is random, and the robot needs to reach the target point within the maximum number of steps allowed in each round. The experimental environment is shown in Table 1, and the experimental parameter information is shown in Table 2.

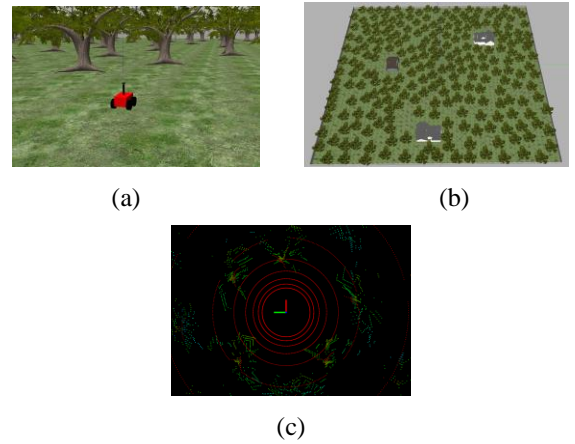

(a)    (b)

(c)

Fig 2. Simulation Environment

TABLE I. EXPERIMENTAL ENVIRONMENT

| Platform | Name |
|---|---|
| Operating system | ROS in Linux |
| Physical simulation environment | Gazebo |
| Data Visualization Tools | Rviz |
| Device hardware configuration | NVIDIAGeForce GTX 1650 Ti/PCLe/SSE2 |

TABLE II. TRAINING PARAMETERS

| Parameter | Numerical value |
|---|---|
| $\varepsilon$-greedy $\varepsilon$ | 0.05 |
| Discount rate $\gamma$ | 0.9 |
| Learning rate $\alpha$ | 0.01 |
| Batch size | 64 |
| Number of iterations | 1500 |
| Replay buffer size | 5000 |

In this section, to verify the performance of the proposed DDPG algorithm, we compared it with DQN and DDQN on a comprehensive test problem.

We constructed a test environment with a rectangular area of 150m × 150m in size, as shown in Figure 2, which contains three houses and several trees evenly distributed throughout the map to test the robot's autonomous exploration ability in an unfamiliar environment.

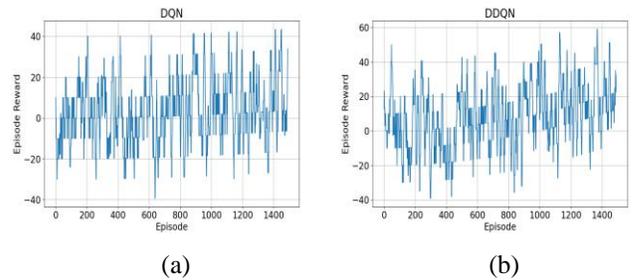

(a)    (b)

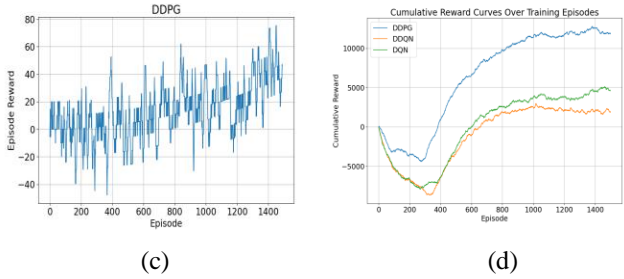

(c)                (d)

Fig 3. Training rewards

According to the results in Figure 3, the larger the reward value, the closer the robot is to the target and the greater the success rate of reaching the target point. As shown in Figure 3(a), the DQN algorithm was trained in a static obstacle environment for 330 episodes, and the reward values at the end of training were still mostly negative; as shown in Figure 3(b), the cumulative reward value obtained by the DDQN algorithm did not show an obvious upward trend at the end of 400 episodes, and the cumulative reward value was still negative at the end of training. As shown in Figure 3(c), during the path planning training of the DDPG algorithm, most episodes after 250 episodes took positive values. Figure 3(d) shows the cumulative reward.

TABLE III. THE PERFORMANCE OF ALGORITHM

| Method | Success rate% | Training model time(h) | Average time(s) |
|---|---|---|---|
| DDPG | 97 | 16 | 45.3 |
| DDQN | 81 | 14 | 62.4 |
| DQN | 76 | 17 | 66.1 |

Table III presents the performance results of three algorithms DDPG, DDQN, and DQN used for training path planning in an autonomous navigation system. The DDPG algorithm achieved the highest success rate at 97% and the shortest average time per episode at 45.3(s), with a training time of 16 hours. The DDQN algorithm showed a success rate of 81%, an average time per episode of 62.4(s), and a training time of 14 hours. The DQN algorithm had the lowest success rate at 76%, the longest average time per episode at 66.1(s), and a training time of 17 hours. These results indicate that the DDPG algorithm is the most effective and efficient among the three.

## IV. CONCLUSIONS

To address the challenges of autonomous navigation of unmanned vehicles, traditional methods often have shortcomings, such as low path planning efficiency, unsatisfactory path results, easy to fall into local optimality, and difficulty in selecting objective function weight coefficients. This paper adopts the DDPG algorithm to improve the real-time planning ability of unmanned vehicles in unfamiliar environments. In a simulated environment, the effectiveness of the method is demonstrated through various comparative experiments, in which DDPG is compared with DQN and DDQN algorithms, showing excellent performance in dynamic and complex scenes.